\pdfoutput=1
\documentclass[11pt]{article}

\usepackage{EMNLP2022}

\usepackage{times}
\usepackage{latexsym}
\usepackage{xcolor}
\usepackage{graphicx}
\usepackage[T1]{fontenc}
\usepackage[utf8]{inputenc}

\usepackage{microtype}

\usepackage{inconsolata}

\usepackage{multicol}
\usepackage{multirow}
\usepackage{booktabs}
\usepackage{colortbl}
\usepackage[ruled, lined, linesnumbered, commentsnumbered, longend]{algorithm2e}
\usepackage{amsmath}
\usepackage{float}

\title{Who Says Elephants Can't Run: \\ Bringing Large Scale MoE Models into Cloud Scale Production}
\author{Young Jin Kim \\
  Microsoft \\ %/ Address line 1 \\
  \texttt{youki@microsoft.com} \\\And
  Rawn Henry \\
  NVIDIA \\ % / Address line 1 \\
  \texttt{rhenry@nvidia.com} \\
  \AND
  Raffy Fahim \\
  Microsoft \\ % / Address line 1 \\
  \texttt{raffybekheit@microsoft.com} \\\And
  Hany Hassan Awadalla \\
  Microsoft \\ % / Address line 1 \\
  \texttt{hanyh@microsoft.com} \\}

\begin{document}
\maketitle
\begin{abstract}
%\textcolor{red}{Rewrite the abstract based on the latest contents}
% TODO abstract need to be revised
Mixture of Experts (MoE) models with conditional execution of sparsely activated layers have enabled training models with a much larger number of parameters. As a result, these models have achieved significantly better quality on various natural language processing tasks including machine translation. However, it remains challenging to deploy such models in real-life scenarios due to the large memory requirements and inefficient inference. In this work, we introduce a highly efficient inference framework with several optimization  approaches to accelerate the computation of sparse models and cut down the memory consumption significantly. While we achieve up to 26x speed-up in terms of throughput, we also reduce the model size almost to one eighth of the original 32-bit float model by quantizing expert weights into 4-bit integers. As a result, we are able to deploy 136x larger models with 27\% less cost and significantly better quality compared to the existing solutions.  This enables a paradigm shift in deploying large scale multilingual MoE transformers  models replacing the traditional practice  of distilling teacher models into dozens of  smaller models per language or task. 
 
\end{abstract}
% several ways to effectively productize large scale MoE models into the cloud scale machine 
%Especially, the proposed 4-bit and 8-bit GEMMs (GEneral Matrix Multiply) can consume 4/8 bit quantized weights without any special instructions, so it can be easily used for the place of normal feedforward layers without having additional logic to handle the precision of activations.

\section{Introduction}
Transformer models are getting larger and better on a continuous basis. The largest transformer models scale up to hundreds of billions of parameters, \cite{megatron-turing540b} resulting in high training and inference costs. This makes it difficult to deploy such models in any real-life scenario with reasonable latency and throughput. Mixture of Experts (MoE) models offer a more cost-effective method to scaling model sizes by using sparsely activated computations. More specifically, feed forward layers can be easily enlarged by replicating the original weights $E$ times where $E$ is the number of experts. Each of these replicas is referred to as an expert, and tokens get routed to these experts depending on a gating function. Transformer models have a much larger number of parameters when utilizing these MoE layers. However, the number of flops remains comparable to their dense counterparts thanks to sub-linear scaling in computation costs ~\cite{shazeer2017outrageously}. 
Recently, the Mixture of Experts (MoE) architecture has been successfully utilized to scale massive large scale multilingual models \cite{lepikhin2020gshard}), NLU tasks \cite{fedus2021switch, st-moe2022} and multilingual multitask models \cite{kim2021scalable}.  

MoE offers the benefits of scaling the model to gain better accuracy without paying the huge compute cost  of massive dense models. However, large scale MoE models bring  their own set of unique challenges to get efficient training and inference methods.  Most of the previous work focused on improving training efficiency and throughput \cite{fedus2021switch, kim2021scalable}. In this work, we focus on optimizing MoE models inference and latency since it is crucial to harvest the benefits of such  models in real-life scenarios. 

\emph{Production-scale Multilingual Machine Translation systems:} 
 in this work, we explore deploying MoE models for large scale Multilingual Machine Translation systems to benefit from large language models, while maintaining reasonable serving cost. Multilingual large scale systems are already very attractive due to multiple aspects. First, they benefit modeling since they allow better accuracy, especially through transfer learning across languages. Additionally, they improve deployment and serving since we can replace dozens of models with a single model that is able to serve many languages at the same time. Nevertheless, we need the inference to be highly optimized to make inference cost-efficient. Despite these benefits, shipping such multilingual models brings a new challenge, because they usually require a much larger model capacity in terms of the number of parameters and the computation. The MoE model architecture could be a promising solution given its sub-linear or constant FLOPs increase in terms of the number of model parameters. But, the large memory consumption issue still remains.

%The current Machine Translation deployment paradigm generally  follows the teacher-student model. Where several teachers are being distilled into very small student model that get deployed on CPU \cite{kim2019research}. The existing approach to deploy more than one hundred different languages into translation services is to train each individual language pair separately. This is not scalable especially each individual model needs to go through various model compression steps to be deployed on CPUs which have relatively low FLOPs numbers. This not only hinders scalable model building, but also knowledge sharing and transfer between different language pairs and tasks. Multilingual and multitask training approaches have been utilized to overcome this problem. However, shipping these multilingual and multitask models brings a new challenge because these models usually require much larger model capacity in terms of the number of parameters and the computation. The MoE model architecture could be a promising solution given its sub-linear or constant FLOPs increase in terms of the number of model parameters. But, the large memory consumption issue still remains.
  
In this work, we show how to enable deploying a single MoE model that can serve many languages replacing dozens of traditional models while improving accuracy and maintaining latency, throughput and cost efficiency. We set the goal for this work to match latency and throughput of a distilled small model deployed on CPU while achieving better serving cost.

It is worth noting that while  the optimizations  presented here are applied to MoE encoder-decoder architecture for multilingual machine translation task, they are applicable to other architectures and tasks without any loss of generality. Given the recent success of MoE models on wide set of NLU  and NLG tasks \cite{fedus2021switch, st-moe2022}, we believe the optimization presented in this work will be equally enabling to other tasks as it is for machine translation. 

 \section{Challenges and Contributions}
%  \section{MoE Inference Optimization}
 
 \subsection{MoE Inference challenge}
 \label{sec:challenges}
 
Even though the MoE architecture in theory requires much less computation with larger number of parameters, it adds several computations such as token routing and all-to-all communication which could be a  significant hit to the training throughput as much as 12\%  for a single node  as shown in~\cite{liu2022gating}. In addition, it significantly increases the amount of memory traffic in the MoE layers. So far, previous studies focused more on the training efficiency of those MoE models and there has not been a solution to deploy this kind of models into the real-time applications. At inference time, we have observed the naive implementation of MoE models could be up to 30 times slower than its dense counterpart with the same embedding and hidden dimensions. To achieve a reasonable deployment cost, it is critical to lower the inference cost by increasing throughput and reducing the latency. Since MoE layers are not widely optimized for the inference scenarios, it is challenging to build efficient runtime environment in terms of computation and memory consumption.

Recently, \cite{ds_moe-infer} introduced several approaches to improve inference of MoE models focusing on very large scale models larger than 100B parameters  and decoding on multiple GPUs. When the model size increases beyond the memory limit of a single  GPU, multiple GPUs can be used together for a single inference by splitting the model weights across different GPUs. While multi-gpu can reduce latency and is required to serve extremely large models, it introduces significant   communication overhead and makes it more difficult to scale up and down the number of instances based on traffic. Therefore, even though multiple GPUs could bring much larger models into production, we focus on the single GPU inference scenario due to its cost efficiency  with reasonably sized models. It is worth noting that the optimization we are presenting here for single GPU can be utilized for larger models on several GPUs as well. However, this is beyond the scope of this paper.

\subsection{Inference Optimization Contributions}
% \textcolor{red}{YOUNG: We need to be clear here on list of concise and clear contributions}
% In this paper, we showcase how we overcame the challenges described in \ref{sec:challenges} and deploy large scale MoE models into production.
In this paper, we show how to reduce the memory requirements to deploy largest possible model on a single GPU, which avoids costly all-to-all collectives. In addition, we optimized  routing efficiency for GPUs and implemented batch pruning. We describe how we extend NVIDIA's FasterTransformer\footnote{\url{https://github.com/NVIDIA/FasterTransformer}} inference framework to support the MoE model architecture in a real world deployment scenario:
\begin{itemize}
    \item We present how we utilize the parallel primitives in the CUTLASS\footnote{\url{{https://github.com/NVIDIA/cutlass}}} and CUB \footnote{\url{{https://github.com/NVIDIA/cub}}} libraries to efficiently express token routing and the batched matrix multiply required for MoE.
    
    %to implement highly optimized parallel algorithms for exploiting massive CUDA cores for the additional operations used in the MoE architecture such as token routing and expert layers.
    
    \item We propose a new GEMM (GEneral Matrix Multiply) which can consume 4-bit/8-bit quantized weights and perform float math. The new GEMM works as drop-in replacements of normal feedforward layers without having additional logic to handle quantization/dequantization of activations. We also show that 4/8 bit weight-only quantization preserves the accuracy without any additional algorithms.

    \item We implement an effective batch pruning algorithm for MoE layers to make the search algorithm on the decoder very efficient.
    % \item We present a highly efficient inference solution on GPU to deploy large MoE models into a cloud scale production.In this section, we describe the extensions made to NVIDIA's FasterTransformer library to efficiently support MoE inference. 
\end{itemize}

\subsection{FasterTransformer overview}
We build our MoE optimization over NVIDIA's  FasterTransfomer, a highly optimized open source  inference  engine for transformer models. FasterTransformer implements a highly optimized transformer layers for both the encoder and decoder for inference which is built on top of CUDA, cuBLAS, cuBLASLt and C++. FasterTransformer supports seamless integration with Triton Inference server \footnote{https://github.com/triton-inference-server/server} which enabled us to deploy our models in scalable  large scale cloud environment. 

We have extended FasterTransformer to support DeepSpeed MoE models\cite{kim2021scalable} and added support for Transformer with Untied Positional Encoding (TUPE)~\cite{ke2020tupe} attention, gate routing and efficient computation of MoE layers, including batch pruning in those layers.

\section{MoE Inference Optimizations}
% \section{Model architecture design}
% \maketitle
\subsection{Model architecture}
MoE showed tremendous success with encoder-decoder model architecture  in Multilingual Machine Translation \cite{lepikhin2020gshard,kim2021scalable},  and in Natural Language understanding \cite{fedus2021switch,st-moe2022}. Therefore, in this work we focus on the encoder-decoder architecture without loss of generality since  the optimization is directly applicable to encoder-only and decoder-only models as well.

We train an encoder-decoder model  for machine translation with deep encoder and shallow decoder architecture  as proposed in ~\cite{kim2019research,kasai2020deep}. 
%, we use a single encoder and a single decoder architecture. 
For a given batch of input sentences, the encoder is executed only once while  the decoder is executed multiple times with a beam search algorithm per token.  The auto-regressive execution of the  decoder is usually the performance bottleneck. Therefore, utilizing a shallow decoder partially  mitigates that effect.  Empirically, we have found that using half number of decoder layers than the number of encoder layers gives a good trade-off between quality and performance.
% We employ MoE architecture to efficiently increase the number of parameters~\cite{kim2021scalable}.
For the most efficient MoE layer execution, we use top-1 gating algorithm proposed in Switch transformers ~\cite{fedus2021switch}. At every other layer, MoE layer is used instead of the plain feedforward layer. %We use a vocabulary consists of 128,000 sub-words to cover multiple different languages in one model.

%\begin{figure}
   
%    \includegraphics[width=7 cm]{images/Moe_Arch.jpg}
%    \caption{Mixture of Experts model architecture. }
%    \label{fig:moe_arch}
%\end{figure}
% \subsubsection{Latency consideration}
% \subsubsection{Model size choices}
\label{ssec:model_sizes}
We use embedding dimension of $1024$, the positional and word correlations are computed separately and added together in the self attention module (TUPE) \cite{ke2020tupe}. The feed-forward hidden dimension is  $4096$ with  $24$ encoder layers and $12$ decoder layers as proposed in \cite{kim2021scalable}. This model configuration satisfies the deep encoder and shallow decoder design and the model weights fit well into the GPU memory without tensor slicing model parallelism~\cite{shazeer2018mesh}. The tensor slicing approach increases communication overheads and could potentially introduce training instability issues. In the production setting, we choose a model building pipeline which could minimize such instability. On the other hand, expert parallelism is preferred over tensor slicing model parallelism because an atomic layer operation such as a feedforward layer is executed inside one GPU. Therefore, we increase the number of model parameters by adding more experts. With the size of the layers and the number of layers, the total number of parameters is roughly $5$ billion when $32$ experts are used in the MoE layers. With half precision floating point (fp16), this is about $10$ GB which can fit on a single $16$ GB GPU.

% - Tradeoffs between model params and constraints related to prod system
% - FLOPs x model params => Connecing FLOPs with model latency
% \subsection{Training data preparation}

\subsection{Multilingual Machine Translation Model}
The traditional  Machine Translation deployment paradigm generally  follows the teacher-student model. Where several teachers are being distilled into a very small student model that get deployed on CPU \cite{kim2019research}. For instance, deploying 100 languages translation system, would require  training, distilling and deploying at least 200 of such models. Each model is trained individually  for a particular language pair. This is not scalable  since  each individual model needs to go through various model compression steps to be deployed on CPUs with relatively low FLOPs numbers. This not only hinders scalable model building, but also knowledge sharing and transfer between different language pairs and tasks. Multilingual  training approaches have been utilized to overcome this problem. However, shipping these multilingual models brings a new challenge since such models usually require much larger capacity in terms of the number of parameters and the computation. 

In this work, we use  a multilingual MT system trained on 10 language pairs and can be used in place of individual systems per language  pair.
%or up to 100 systems taking into account direct translation between any two languages.  
The model is trained  using production  scale training data of up to $\sim$ 4B training  sentence  pairs with a vocabulary of 128K using Sentence Piece \footnote{https://github.com/google/sentencepiece}

\subsection{Optimized GPU kernel design}
\label{section:opt_kernel}
One key factor to get an optimal performance with massive CUDA cores is to have efficient parallel algorithms for various additional operations for MoE. In MoE layers, each row in the input activation must get routed to a specific expert weight matrix, depending on a top-$k$ gating function. We implement this routing as a GPU friendly radix sort using NVIDIA's highly efficient CUB library.

In this case, each row in the activation matrix is a token to be translated. The top-$k$ gating function outputs a list with $k$ $(expert\_scale, expert\_idx)$ tuples for each input token. Thus, for top-1 gating (as is done in our case), the function outputs a single tuple for every row of the activation matrix.

In order to perform the routing, we first append the index for each row to the end of the tuple giving a tuple of $(expert\_scale, expert\_idx, row\_idx)$. Then, we sort the tuple using $expert\_idx$ as the keys in order to group all rows that will be processed by the same $expert\_idx$ together. The $row\_idx$ entry from the sorted tuples are then used to permute the original activation matrix in global memory to a layout where all rows routed to the same expert are laid out contiguously in memory.

\begin{figure}
    % \centering
    \includegraphics[width=\columnwidth]{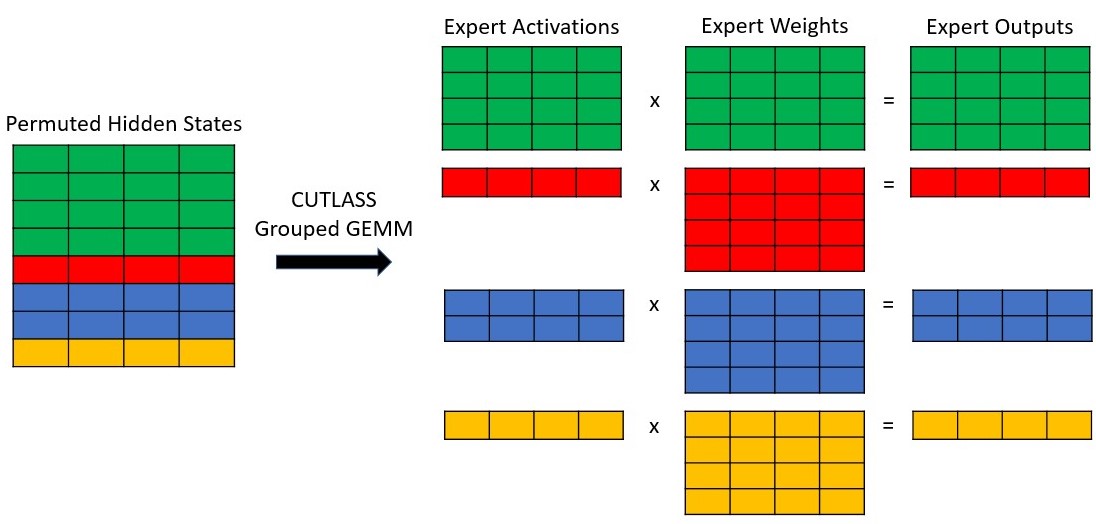}
    \caption{Shows the computation performed by CUTLASS Grouped GEMM. Each color is a sub-matrix for a particular expert, with the matrix multiplies for each expert happening in parallel. If the yellow sentence was finished, it would be omitted from the computation with batch-pruning enabled. This would completely remove the need to load the weight matrix for the yellow expert.}
    \label{fig:moe_gemms}
\end{figure}

In order to finalize the routing, we view each group of rows assigned to a particular expert as its own sub-matrix and compute pointers to the start of these sub-matrices. We then pair each sub-matrix pointer with pointers to the weights and biases for the expert they are routed to, and use CUTLASS Grouped GEMM to compute all of these matrix multiplies in parallel using a single kernel. Figure \ref{fig:moe_gemms} shows the computation performed by CUTLASS.

Finally, we un-permute the rows to their original ordering and apply the expert\_scale to each row before passing the output of the MoE module to the other parts of the network.

\subsection{Expert quantization with 4-bit and 8-bit}
\label{section:quantization}

We quantize the MoE weights for two reasons:
\begin{enumerate}
    \item MoE weights are extremely large which limits the size of the models that can fit on the common 16 GB inference cards such as T4.

    \item MoE matrix multiplies require loading the weights for several different experts which results in them being memory bound.
\end{enumerate}

We do not use Quantization Aware Training (QAT)~\cite{wu2020integer}, because our quantization approach does not degrade model performance. QAT is usually used when there exists a noticeable performance degradation from quantization. Also, we focus on quantizing expert weights only, because they are contributing to more than 90\% of entire model weights thanks to the special property of MoE model size scaling. We get much larger model mostly from the expert parameters in MoE layers ~\cite{shazeer2017outrageously}.

\begin{algorithm}
    \DontPrintSemicolon
    \footnotesize
    \caption{Weight dequantize}\label{alg:dequantize}
    \SetKwFunction{NewMatrix}{NewMatrix}
    \SetKwFunction{IntToFloat}{IntToFloat}
    % \SetKwInput{Input}{Input}
    % \SetKwInput{Output}{Output}
    \SetKwInOut{KwIn}{Input}
    \SetKwInOut{KwOut}{Output}
    
    \KwIn{$E$ - Number of Experts}
    \KwIn{$W$ - quantized weights of shape $(E, M, N)$}
    \KwIn{$S$ - FP16 scales of shape $(E, 1, N)$}

    \KwOut{FP16 dequantized weights}
    \BlankLine
    $W_{dq} \leftarrow \NewMatrix(E, M, N)$

    % \tcc{For odd elements in the list, we add 1, and for even elements, we add 2.
    % After the loop, all elements are even.}
    \For{$e \leftarrow 0$ \KwTo $E-1$}{
            \For{$m \leftarrow 0$ \KwTo $M-1$}{
                    \For{$n \leftarrow 0$ \KwTo $N-1$}{
                        $f = \IntToFloat(W[e, m, n])$ \;
                        $W_{dq}[e, m, n] = f * S[e, n]$ \;
            }
        }
    }
    \KwRet{$W_{dq}$}
\end{algorithm}

All activations and biases are kept as FP16 and only the expert weight matrices are quantized. As a result, we do not require any post-training calibration (because we don't need scales for the activations) which makes this recipe easy to apply to several language families. We perform symmetric, range-based per-channel quantization on each expert weight. This means that for expert weights of shape $(E, M, N)$ where $E$ is the number of experts and $M$ and $N$ are arbitrary dimensions, we produce scales of shape $(E, 1, N)$. The same quantization method is used for int4 and int8. During inference, we dequantize the weights to FP16 and perform our matrix multiplies using floating point computations. Algorithm \ref{alg:dequantize} shows the dequantization performed during inference.

One option for implementing the GEMM + Dequantize would be to write a separate kernel to dequantize the weights before the MoE GEMM. However, this would actually increase the amount of memory traffic as we would add a read of $W$ and a write to $W_{dq}$ as shown in Algorithm \ref{alg:dequantize}. As a result, we decided to take advantage of the flexibility of CUTLASS and fuse the dequantize step into the GEMM kernel. After profiling, we realized that the conversion from int to float (line 5 in Algorithm \ref{alg:dequantize}) was slower than anticipated. In order to improve this, we replaced the native int to float conversion (I2F) with a series of high throughput ALU and FP16 instructions which improved the performance of our fused GEMM + Dequantize.

\subsubsection{Quantization Optimization}
The conversion optimization mentioned above produces exact results to the native I2F conversions. It relies on two key observations.

\begin{enumerate}
    \item For any FP16 number $X$ where $1024 \leq X < 2048$, 1024 will be represented exactly in the exponent bits and $int(X - 1024)$ will be directly stored in the mantissa. For example, FP16 representation of 1027 (represented as 0x6403) has the integer 3 stored directly in the mantissa bits of its representation.
    \item For any integer $0 \leq Y < 1024$, we can construct the FP16 representation of $Y + 1024$ by setting the exponent to 1024 and storing $Y$ in the FP16 mantissa. This is easily done by performing 0x6400 | $Y$, since 0x6400 is the hex representation of 1024 in FP16.
\end{enumerate}

Our optimization exploits these observations to quickly convert int4s or int8s and FP16. After we quantize the weights, we add 128 to int8 weights and 8 to int4 weights to make them all unsigned. We refer to these weights as $W_{+}$. This is not strictly necessary, but removes the need to perform sign extension logic.

\subsubsection{Optimized 8-bit Dequantize}
In order to best utilize the hardware, we convert int8s to FP16s two at a time, leveraging the fact that 2 FP16 elements can fit in a 32-bit register. This is done as follows:

\begin{enumerate}
    \item We load 4 int8 values, $[e_0$, $e_1$, $e_2$, $e_3]$ from $W_+$ into a single 32-bit register.
    \item We then create a second 32-bit register, $R_1$, that stores the FP16 representation of $[e_0 + 1024, e_1 + 1024]$  leveraging observation (2).
    \item Next, we use float math to subtract $[1152, 1152]$ from $R_1$. This subtraction is due to the fact that we must subtract 1024 from each number in $R_1$ convert $e_0$ and $e_1$ to FP16. Then, we must further subtract 128 from each number to obtain the float representation of the original, signed integer.
    \item Lastly, we repeat steps 2 and 3 for $e_2$ and $e_3$.
\end{enumerate}

\subsubsection{Optimized 4-bit Dequantize}
We change the layout of the weights to reduce the number of logic instructions needed to construct the FP16s $[e_i + 1024, e_{i+1} + 1024]$ . Thus, for int4, we change the layout of $W_+$ to reorder groups of 8 elements as follows:
\begin{equation*}
\resizebox{\columnwidth}{!}{$[e_0, e_1, e_2, e_3, e_4, e_5, e_6, e_7 ] \rightarrow [e_0, e_2, e_4, e_6, e_1, e_3, e_5, e_7]$}
\end{equation*}
With this new layout, the idea for int4 is similar to what was previously described for int8. Of course, we must now subtract $[1032, 1032]$ to recover the original, signed integer as fp16. We must also iterate 4 times since 1 32-bit register holds 8 int4s and conversion happens 2 at a time. 

\subsection{MoE Batch Pruning}
\label{section:batch_pruning}
% Auto regressive decoders translate sentences one token at a time over a series of time steps. They produce $beam\_width$ candidate output tokens at every time step for each sentence in a batch. Every sentence will produce a different number of output tokens. This causes different sentences in a batch to complete translation at different time steps. As a result, there can be wasted computation if we keep processing finished sentences without removing them from the batch. 

Batch pruning refers to the act of removing sentences from a batch dynamically as soon as they are done translating. We observed that this speeds up MoE layers as it can prevent the loading of entire expert weights, reducing the amount of memory traffic required in these memory bound layers. 

In order to implement batch pruning in the MoE layers, we make a simple modification to the gating function so that it assigns a large expert\_idx to all finished sentences. This causes all finished sentences to be moved to the end of the permuted activation matrix in the routing step. To complete the pruning, we simply keep track of the total number of active tokens and only process the first active\_tokens rows of the permuted activation matrix mentioned in section \ref{section:opt_kernel}.

% \textbf {Results}

% - See results sections :D

% \textbf {Future Work or Limitations}

% - First implementation does not fully vectorize the global memory loads for the weight matrices. Doing this much more complicated and requires changes deeper in CUTLASS. It is a future task. Specifically, always loading 8 elements (regardless of datatype) allows us to reuse all of the CUTLASS infra for fast gemm, including SMEM layouts.

% \textcolor{blue}{4bit quantization single GEMM comparison: Batch size, num experts, quantization }
% - MoE specific: memory bound (loading the weights)
% - Doesn't require activation scaling
% - Using fp16 instructions can work for all tensorcore architectures

% \textcolor{green}{Reuse of CUTLASS efficient GEMM pipeline. Working mixed type gemm with small additions.}

% \textcolor{green}{Currently loads don't make best use of DRAM (not fully vectorized) so impl is not optimal.}
% \textcolor{green}{Can improve via shared memory, but did not have time. Just say that it is complicated due to SMEM layouts \& we did not have time to implement, but plan to in the future.}

\section{Results and discussion}

All experiments in this section are run on a single NVIDIA PCIE V100 running inside a docker container running Ubuntu 20.04 and CUDA 11.6. All code is compiled with nvcc and gcc/g++ 9.3. 

We run our experiments considering an encoder-decoder MoE model with $32$ experts with TUPE ~\cite{ke2020tupe}, similar to the setup in \cite{kim2021scalable} but with a vocabulary size of $128$k. All throughput metrics measure the time to translate 1000 tokenized English sentences ($\sim$ 40K tokens) to German (en-de) or vice-versa (de-en) and record the total number of input tokens translated per second. BLEU metrics are reported on the same data set.

\subsection{Speed-up and Cost-Effectiveness}

We measure the improvement of our batch pruning optimization by comparing the throughput with and without that optimization. We found that we achieve up to $1.14 \times$ speed up relative to our optimized baseline without batch pruning.

\begin{table}[t]

  \footnotesize
  \caption{Throughput of quanitzed MoE GEMMs normalized against the throughput of the FP16 MoE Gemm. The number of active experts is the number of experts that receive tokens from routing. The matrix shapes for the GEMM C = A @ B are A=$m$x1024, B=1024x4096, where $m$ is different for each expert. The total number of tokens is set to 40 since this is close to what the decoder computes in our inference environment.}
  \centering
  {\small
  \definecolor{Gray}{gray}{0.85}
  \resizebox{\columnwidth}{!}{%
  \begin{tabular}{ccccc}
    \toprule
Active Experts & FP16 & Int8 native I2F & Int8 optimized I2F & Int4 optimized I2F \\
\midrule
1 & 1 & 1.05 & 1.28 & 1.24   \\
    \midrule
4 & 1 & 1.01 & 1.21 & 1.28   \\
    \midrule
8 & 1 & 1.34 & 1.21 & 1.57   \\
    \midrule
16 & 1 & 1.40  & 1.39 & 1.73  \\
    \midrule
24 & 1 & 1.40  & 1.49 & 1.78   \\
    \midrule
32 & 1 & 1.46 & 1.59 & 1.85  \\
    \midrule
GEOMEAN & 1 & 1.26 & 1.35 & 1.56\\
    \bottomrule
  \end{tabular}%
  }
  }
  \label{table:i2f_comparison}
\end{table}

\textbf{INT8/INT4 GEMM Performance.}
First, Table \ref{table:i2f_comparison} shows a performance comparison for the FP16 GEMM compared to fused GEMM + Dequantize with native I2F and our optimized I2F sequence for INT8. Our INT4 implementation only supports the optimized I2F sequence. Depending on the number of experts, INT8 and INT4 could accelerate MoE computation up to 59\% and 85\%, respectively.

%We also consider the impact of INT8 and INT4 expert quantization on BLEU scores as we expect translation quality degradation to occur when quantizing model weights ~\cite{kim2019research}. Table \ref{table:bleu_delta} shows the change in BLEU compared to FP16 after applying quantization.
\textbf{INT8/INT4 Quality Impact.}
We also consider the impact of INT8 and INT4 expert quantization on BLEU scores,   we observe  negligible translation quality degradation when quantizing model weights. Table \ref{table:bleu_delta} shows the change in BLEU compared to FP16 after applying quantization.

\textbf{End-to-end Performance Improvements.}
Table~\ref{table:moe_performance} shows our machine translation experiments for EN-DE, with different batch sizes and different quantization schemes and reports both the throughput of our PyTorch and Faster Transformer implementations. Compared to the Torch-FP16 baselines, the optimizations applied achieve significant speed-up across different settings.

\textbf{Cost Comparison.}
Table \ref{table:moe_cost} shows the deployment cost comparison between the MoE models and smaller models optimized for CPU deployment \cite{kim2019research}. The cost of deploying MoE models which are 136x larger on CPU is more than 100 times of the cost of deploying smaller models on CPU. However, the optimized large MoE models on GPU cost less than the current CPU model deployment with smaller models.

% To validate that our system could satisfy our production latency requirements, we measure the inference throughput speed-up on a single GPU compared to our original model running in PyTorch without any of the proposed optimizations. 

\begin{table}[!htbp]
  \footnotesize
  \caption{BLEU differences from INT8 and INT4 weight-only compared to the FP16 baseline. }
  \label{table:bleu_delta}
  \centering
  {\footnotesize
  \definecolor{Gray}{gray}{0.85}
  \setlength{\tabcolsep}{4pt}
  \resizebox{\columnwidth}{!}{%
  \begin{tabular}{ccc}
    \toprule
    \multicolumn{1}{c}{\multirow{3}{*}{Language Pair}} & \multicolumn{2}{c}{Beam 1 $\Delta$ BLEU
    }  \\
    \cmidrule(lr){2-3} %\cmidrule(lr){4-5} 
 & \multicolumn{1}{c}{INT8} & \multicolumn{1}{c}{INT4} \\ %& \multicolumn{1}{c}{INT8} & \multicolumn{1}{c}{INT4} \\
    \midrule
 EN-DE (Beam 1) & -0.028 & -0.052 \\
    \midrule
 EN-DE (Beam 2) & 0.051  & -0.180 \\
     \midrule
 DE-EN (Beam 1) & -0.084 & 0.044 \\
    \midrule
 DE-EN (Beam 2) & -0.027 & -0.031 \\
     \midrule
 Avg. of 10 language pairs (Beam 2) & -0.007 & -0.167 \\
    \bottomrule
  \end{tabular}%
  }
  }
\end{table}

\begin{table*}[!htbp]
  \small
  \caption{Throughputs for beam=1 and beam=2 for varying batch sizes. Throughput is measured as input tokens processed per second. The precisions (FT-INT8 and FT-INT4) in the table refer to the quantization applied to the MoE weights. \textit{Torch-FP16} columns show the throughput numbers when we run the model with PyTorch v1.10 using FP16 model weights.}
%   \caption{Normalized throughputs for beam 1 and beam 2 for varying batch sizes. Throughput is measured as input tokens processed per second and is normalized against a baseline PyTorch implementation. The precisions (FT-INT8 and FT-INT4) in the table refer to the quantization applied to the MoE weights. }
  \label{table:moe_performance}
  \centering
  {\small
  \definecolor{Gray}{gray}{0.85}
  \setlength{\tabcolsep}{4pt}
  % \resizebox{\columnwidth}{!}{%
  \begin{tabular}{ccccc|cccc}
    \toprule
    \multicolumn{1}{c}{\multirow{3}{*}{Batch Size}} & \multicolumn{4}{c}{Beam=1 Input tokens processed/sec
    } & \multicolumn{4}{c}{Beam=2 Input tokens processed/sec} \\
    \cmidrule(lr){2-5} \cmidrule(lr){6-9} 
& \multicolumn{1}{c}{Torch-FP16} & \multicolumn{1}{c}{FT-FP16} & \multicolumn{1}{c}{FT-INT8} & \multicolumn{1}{c}{FT-INT4} &  \multicolumn{1}{c}{Torch-FP16} & \multicolumn{1}{c}{FT-FP16} & \multicolumn{1}{c}{FT-INT8} & \multicolumn{1}{c}{FT-INT4} \\
    \midrule
1 & 16 & 388 & 401 & 400 & 14 & 351 & 361 & 361  \\
    \midrule
8 & 70 & 1594  & 1639 & 1662 & 65 & 1453 & 1507 & 1518  \\
    \midrule
20 & 150 & 3025 & 3178 & 3247 & 139 & 2571 & 2719 & 2803 \\
    \midrule
32 & 214 & 4008 & 4264& 4379 & 202 & 2960 & 3137 & 3239  \\
    \midrule
64 & 379 & 5371 & 5706 & 5935 & 349 & 4333 & 4578 & 4746 \\
    \midrule
96 & 485 & 6689 & 7101 & 7483 & 440 & 5062 & 5384 & 5605 \\
    \bottomrule
  \end{tabular}%
  }
  %}
\end{table*}

\begin{table*}[!htbp]
  \small
  \caption{Deployment cost comparison. We show the most cost-effective throughputs under our 1s latency budget.}
  \centering
  {\small
  \definecolor{Gray}{gray}{0.85}
  \setlength{\tabcolsep}{4pt}
%   \resizebox{\columnwidth}{!}{%
  \begin{tabular}{c|c|c|c|c|c|c}
    \toprule
    Hardware & Parameters & Batch size & Price (East US) & Latency (ms) & Throughput (words/sec) & Monthly USD/token \\
    \midrule
    CPU (AVX512) & 0.04 B & 1 & \$587.65 (F16s) & 75 & 351 & 0.209 \\
     \midrule
    \midrule
    CPU (AVX512) & 5.32 B & 1 & \$587.65 (F16s) & 1080 & 26 & 22.602 \\
     \midrule
    \multirow{2}{*}{NVIDIA T4} & \multirow{2}{*}{5.32 B} & \multirow{2}{*}{20} & \$390.55  & \multirow{2}{*}{421} & \multirow{2}{*}{1565} & \multirow{2}{*}{0.250} \\
     &  &  &(NC4as T4 v3) &  &  &  \\
     \midrule
    \multirow{2}{*}{NVIDIA T4} & \multirow{2}{*}{5.32 B} & \multirow{2}{*}{64} & \$390.55 & \multirow{2}{*}{824} & \multirow{2}{*}{2560} & \multirow{2}{*}{0.153} \\
     &  &  & (NC4as T4 v3) &  &  &  \\
    \bottomrule
  \end{tabular}%
%   }
  }
  \label{table:moe_cost}
\end{table*}

% \textcolor{red}{we may want to use two column table for a few of the tables.}

% - \textcolor{blue}{How to discuss inference COGS savings \textbf{on T4} (cost-effectiveness)? - being low-cost is important compared to cpu models. "We found it to be the most cost-effective gpu..." "dollar/inference". Need clarification of CPU numbers}

% - \textcolor{blue}{Why we chose V100 gpu over T4: better latency}

% \textcolor{blue}{- Compare training cost between multilingual and bilingual systems}

% - \textcolor{orange}{I think we could also talk about the quantization experiments we ran with 8-bit and 2-bit to say what didn't work}
\section{Conclusions and Future Work}
This paper describes how to make large MoE models cost-efficient on a single GPU in a real-world inference environment. The final implementation achieves a speedup of up to 26X over PyTorch baseline. Our GPU MoE implementation allows serving much larger and higher-quality models compared to dense models on CPUs without increasing the cost of serving. We consider  two main avenues for future work. We are currently working on improving our fused GEMM + Dequantize kernel to enable the use of fully vectorized 16 byte loads on the weight matrix. In addition, we plan to explore deploying even larger models with distributed inference in the future in a cost-efficient way.

\section*{Ethics Statement}
The authors have put the best effort to comply with the \href{https://www.aclweb.org/portal/content/acl-code-ethics}{ACL Ethics Policy}. For the experiments, we have used WMT public domain datasets and respected the license policy for our usage.
% Scientific work published at EMNLP 2022 must comply with the \href{https://www.aclweb.org/portal/content/acl-code-ethics}{ACL Ethics Policy}. We encourage all authors to include an explicit ethics statement on the broader impact of the work, or other ethical considerations after the conclusion but before the references. The ethics statement will not count toward the page limit (8 pages for long, 4 pages for short papers).
\section*{Acknowledgements}
We thank the Microsoft Z-Code team and the Microsoft Translator team for the great effort to push the limit of the production quality in machine translation and to quickly adopt the state-of-the-art Mixture of Experts models into the cloud-scale production. We also thank the Microsoft DeepSpeed team for the collaboration on more efficient and scalable Mixture of Experts architecture and library development. Additionally, we are deeply grateful for the amazing work of the NVIDIA CUTLASS team which developed grouped GEMM kernels which were crucial to get good performance with Mixture of Experts. We also thank the CUTLASS team for answering all of our questions to help us implement efficient kernels that handle GEMMs with different input types. Lastly, we thank the NVIDIA FasterTransformer team for their help with integrating our Mixture of Experts implementation into their efficient transformer inference framework.
% \raggedbottom
% @Rawn - CUTLASS? FT?

% This document has been adapted by Yue Zhang, Ryan Cotterell and Lea Frermann from the style files used for earlier ACL and NAACL proceedings, including those for 
% ACL 2020 by Steven Bethard, Ryan Cotterell and Rui Yan,
% ACL 2019 by Douwe Kiela and Ivan Vuli\'{c},
% NAACL 2019 by Stephanie Lukin and Alla Roskovskaya, 
% ACL 2018 by Shay Cohen, Kevin Gimpel, and Wei Lu, 
% NAACL 2018 by Margaret Mitchell and Stephanie Lukin,
% Bib\TeX{} suggestions for (NA)ACL 2017/2018 from Jason Eisner,
% ACL 2017 by Dan Gildea and Min-Yen Kan, NAACL 2017 by Margaret Mitchell, 
% ACL 2012 by Maggie Li and Michael White, 
% ACL 2010 by Jing-Shin Chang and Philipp Koehn, 
% ACL 2008 by Johanna D. Moore, Simone Teufel, James Allan, and Sadaoki Furui, 
% ACL 2005 by Hwee Tou Ng and Kemal Oflazer, 
% ACL 2002 by Eugene Charniak and Dekang Lin, 
% and earlier ACL and EACL formats written by several people, including
% John Chen, Henry S. Thompson and Donald Walker.
% Additional elements were taken from the formatting instructions of the \emph{International Joint Conference on Artificial Intelligence} and the \emph{Conference on Computer Vision and Pattern Recognition}.

\bibliography{anthology,custom}
\bibliographystyle{acl_natbib}

% \appendix

% \section{Example Appendix}
% \label{sec:appendix}

% This is a section in the appendix.

\end{document}